\documentclass[twoside,11pt]{article}

\usepackage[preprint]{jmlr2e}

\usepackage{amsmath}
\usepackage{fontawesome}
\usepackage{hyperref}
\usepackage{subfig}

\hypersetup{
  colorlinks=true,
  citecolor=black,
  linkcolor=black,
  filecolor=black,
  urlcolor=cyan,
}

\jmlrheading{1}{2021}{1-48}{4/00}{10/00}{kreiss21a}{Sven Kreiss}

\ShortHeadings{Deep Social Force}{Kreiss}
\firstpageno{1}

\begin{document}

\title{Deep Social Force}

\author{\name Sven Kreiss \email research@svenkreiss.com \\
       \addr VITA lab,
       École Polytechnique Fédérale de Lausanne (EPFL)\\
       Lausanne, CH-1015, Switzerland}


\maketitle

\begin{abstract}
  The Social Force model introduced by Helbing and Molnar in 1995
  is a cornerstone of pedestrian simulation. This paper
  introduces a differentiable simulation of the Social Force model
  where the assumptions on the shapes of interaction potentials are relaxed
  with the use of universal function approximators in the form of neural
  networks.
  Classical force-based pedestrian simulations suffer from unnatural
  locking behavior on head-on collision paths. In addition, they cannot
  model the bias
  of pedestrians to avoid each other on the right or left depending on
  the geographic region.
  My experiments with more general interaction potentials show that
  potentials with a sharp tip in the front avoid
  locking. In addition, asymmetric interaction potentials lead to a left or right
  bias when pedestrians avoid each other.
\end{abstract}


\section{Introduction}

The goal is to model pedestrian behavior with a higher fidelity.
So far, pedestrian simulations do not even model people avoiding each other
on the right in continental, western Europe which limits any application of
pedestrian simulations to counter flows: people passing by each other in
corridors. In fact, most force-based pedestrian simulation models display
unnatural behavior when two people need to avoid each other in head-on paths.

This project brings together classical pedestrian simulations,
deep neural networks and
differentiable simulations. However,
this project will not do a sweeping black-box replacement of the entire model.
The goal is to generalize the Social Force model~\citep{helbing1995social}
with deep learning while keeping its
interpretable nature of forces derived from gradients of scalar potentials.

\paragraph{Social Force in Transportation, Robotics and Crowd Modeling.}
The Social Force model is a cornerstone of
pedestrian modeling. It is a micro-simulation of pedestrian behavior based
on simple interaction potentials that exhibit complex behaviors at scale.
Social Forces are the gradients of interpretable potentials.

While the interpretability of the interaction potentials is a feature of the
Social Force model, their shape constraint to a family of falling exponential
functions, or any other parametric form, is unnatural. The model would preserve
its interpretability if the interaction potentials were replaced by arbitrary,
continuous and differentiable functions.

\paragraph{Universal Functions for Interaction Potentials.}
This paper describes interaction potentials by universal function
approximators in the form of deep neural networks. In this form, Social Forces are
derived from gradients of the neural network using automatic differentiation
with efficient BackProp~\citep{lecun2012efficient}. The same automatic
differentiation implementation is used to compute the gradients of these
potentials.

The ultimate goal is to use a more expressive model combined with a stochastic
learning technique to infer model parameters from larger amounts of data. It will
still be possible to plot these potentials and to interpret their shapes
as before.



\begin{figure}
  \centering
    \includegraphics[width=0.35\linewidth]{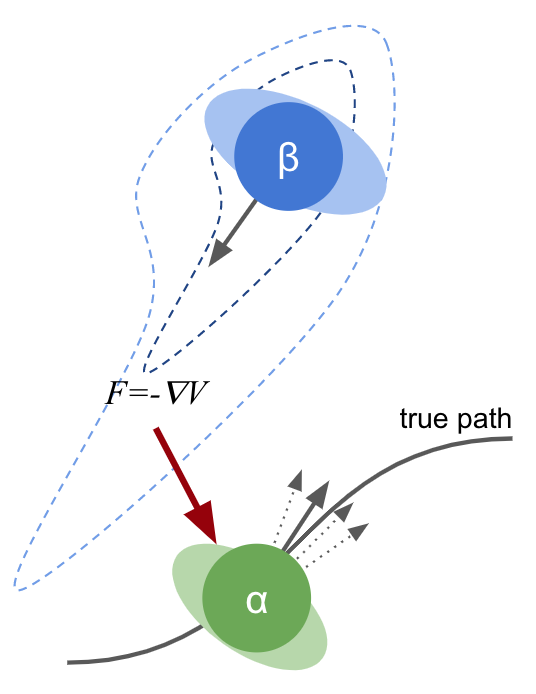}
  \caption{
    Illustration of a pedestrian interaction with the Deep Social Force model.
    In this interaction, pedestrian $\alpha$ feels a force~$F$ that is
    created by the gradient of the
    repulsive potential~$V$ that surrounds pedestrian~$\beta$
    (illustrated by blue, dashed equipotential lines). From
    the predicted path of pedestrian~$\alpha$, an optimizer
    can modify the generic shape of~$V$ so that the prediction will match the
    true path.
  }
  \label{fig:pull-figure}
\end{figure}

This paper proposes a method that retains the structure and interpretability
of the Social Force model and removes assumptions on the shapes of the
interaction potentials. The interaction potentials underlying the Social Force
model are of particular parametric forms. While the
parameters of these potentials can be adjusted to better approximate
real-world data, the
shape is constrained by the choice of the parametric function. Here,
these functions are learned with neural networks without constraining them to
particular shapes. Figure~\ref{fig:pull-figure}, illustrates how different
choices of parameters
for the interaction potential influence the predicted path of a pedestrian.
Social interactions among pedestrians are complex and therefore should not
rely on a particular parametric form.

The software is available at \url{https://github.com/svenkreiss/socialforce}.
All results are produced with an executable book~\citep{executablebookscommunity2020}
hosted at
\url{https://www.svenkreiss.com/socialforce/}. Figures contain deep links
to animated visualizations with this
\href{https://www.svenkreiss.com/socialforce/}{\faVideoCamera~Animation}
symbol or to executable Jupyter notebooks with the
\href{https://www.svenkreiss.com/socialforce/}{\faBook~Notebook} symbol.

\section{Related Work}

This paper focuses on predicting human navigation patterns in crowded scenes where
human-space and more importantly human-human interactions are key to model. As
briefly introduced in the previous
section, there are two general streams of methods: interpretable low capacity ones
and data-driven high capacity ones with low interpretability.

\paragraph{Interpretable models.}

Most of interpretable models are based on hand-crafted energy potentials that
require some domain expertise. Helbing and Molnar~\citep{helbing1995social} have
highly impacted the field by introducing a pedestrian motion model based on
attractive and repulsive forces referred to as the \emph{Social Force} model.
This has been shown to achieve competitive results even on modern pedestrian
datasets~\citep{lerner2007crowds,pellegrini2009you}. This method was used
extensively to study panic behavior in dense
crowds~\citep{helbing2005self, helbing2009pedestrian, helbing2000simulating},
applied in robotics~\citep{luber2010people} and activity
understanding~\citep{mehran2009abnormal,yamaguchi2011you,pellegrini2010improving,leal2011everybody,leal2014learning,choi2012unified,choi2014understanding}.
Variants of the model have been proposed by extending it to a multi-class
case~\citep{robicquet2016learning} or by including rotational
inertia~\citep{headedsocialforce}. Using a dataset of dense crowds,
Seer~\citep{seer2012kinects} have calibrated the Social Force parameters to
produce a model with better predictive power.
Recently, Chen~\citep{chen2018social} presented a nice review on all extensions of the Social Force model.

Related approaches have been used to model human-human interactions.
Treuille~\textit{et. al.}~\citep{treuille2006continuum} use continuum dynamics,
Antonini~\textit{et. al.}~\citep{antonini2006discrete} propose a Discrete Choice
framework and Wang~\textit{et. al.}~\citep{wang2008gaussian},
Tay~\textit{et. al.}~\citep{tay2008modelling} use Gaussian processes.
Such functions have also been used to study stationary
groups~\citep{yi2015understanding,parksocial15}.

Analyzing the potentials behind all these hand-crafted models is in fact the
backbone for interpretability. However, the manually chosen functions behind
these potentials limit the predictability power of these models.

\paragraph{Non-interpretable high capacity models.}

High capacity models learn human-human and human-space interactions in a more generic,
data-driven fashion. They are data hungry with respect to hand-crafted methods.
Yet, with enough data, they outperform previous works in terms of prediction
accuracy. Since 2016, a collection of methods have been proposed based on
Recurrent Neural Networks (RNN)~\citep{pfeiffer2017data, alahi2016social, gupta2018social}.
The challenge remains on how to design the RNN architectures, or how to
represent interactions. Alahi~\textit{et al.}~\citep{alahi2016social} suggest a
social tensor to represent human-human interactions.
Gupta~\textit{et al.}~\citep{gupta2018social} present an adversarial training to learn
the distribution of trajectories. While these methods have the capacity to
learn complex interactions, it is difficult to understand the ``why'' or simply
the impact of surrounding humans on each other. Hence, we propose to take the
benefits of both streams of methods and introduce a non-parametric, Deep Social Force model.

\section{Social Force Examples}

Several toy examples that demonstrate the properties of Social Force
were introduced in the original paper~\citep{helbing1995social}.
These examples are re-implemented here with modern software tools and open sourced.

\begin{figure}
  \centering
  \includegraphics[width=0.8\linewidth,trim=0 0 13cm 0,clip]{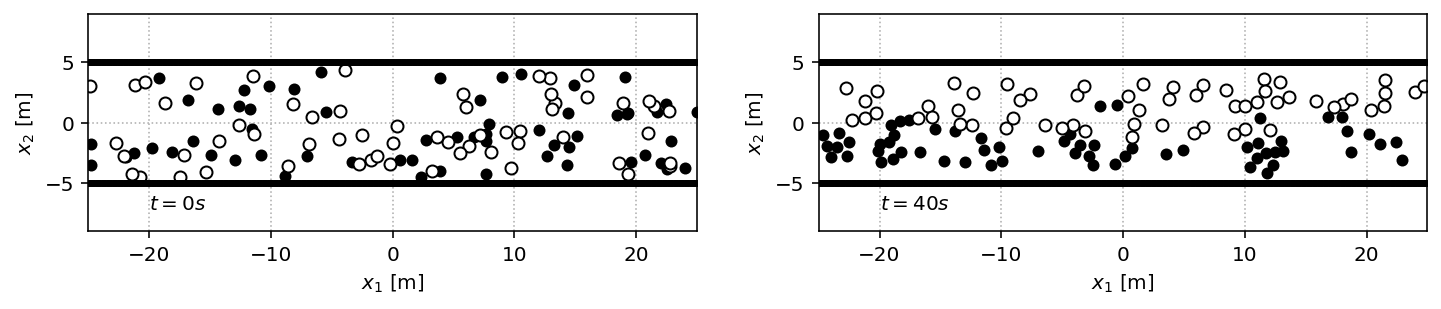}\\
  \caption{%
    Corridor example.
    Pedestrians represented by filled circles walk towards a goal on the right
    and pedestrians represented by empty circles walk towards a goal on the left.
    The size of the circle represents the walking speed of a pedestrian as
    in \cite{helbing1995social}.
    \href{https://www.svenkreiss.com/socialforce/corridor.html\#reference-potential}{\faVideoCamera~Animation}.
  }
  \label{fig:sf-corridor}
\end{figure}

\paragraph{Corridor.}
Figure~\ref{fig:sf-corridor} shows a reproduction of the corridor example.
Pedestrians represented by filled circles walk towards a goal on the right
and pedestrians represented by empty circles walk towards a goal on the left.
The size of the circle represents the walking speed of a pedestrian as
in the original paper.
The original paper included a study on lane-forming behavior that is
not reproduced here. However, once we have generalized interaction
potentials, an extension of this example will show the effect of left-right
asymmetries in interaction potentials in Section~\ref{sec:results}.

\begin{figure}
  \centering
  \includegraphics[width=0.6\linewidth,trim=0 0 0 0,clip]{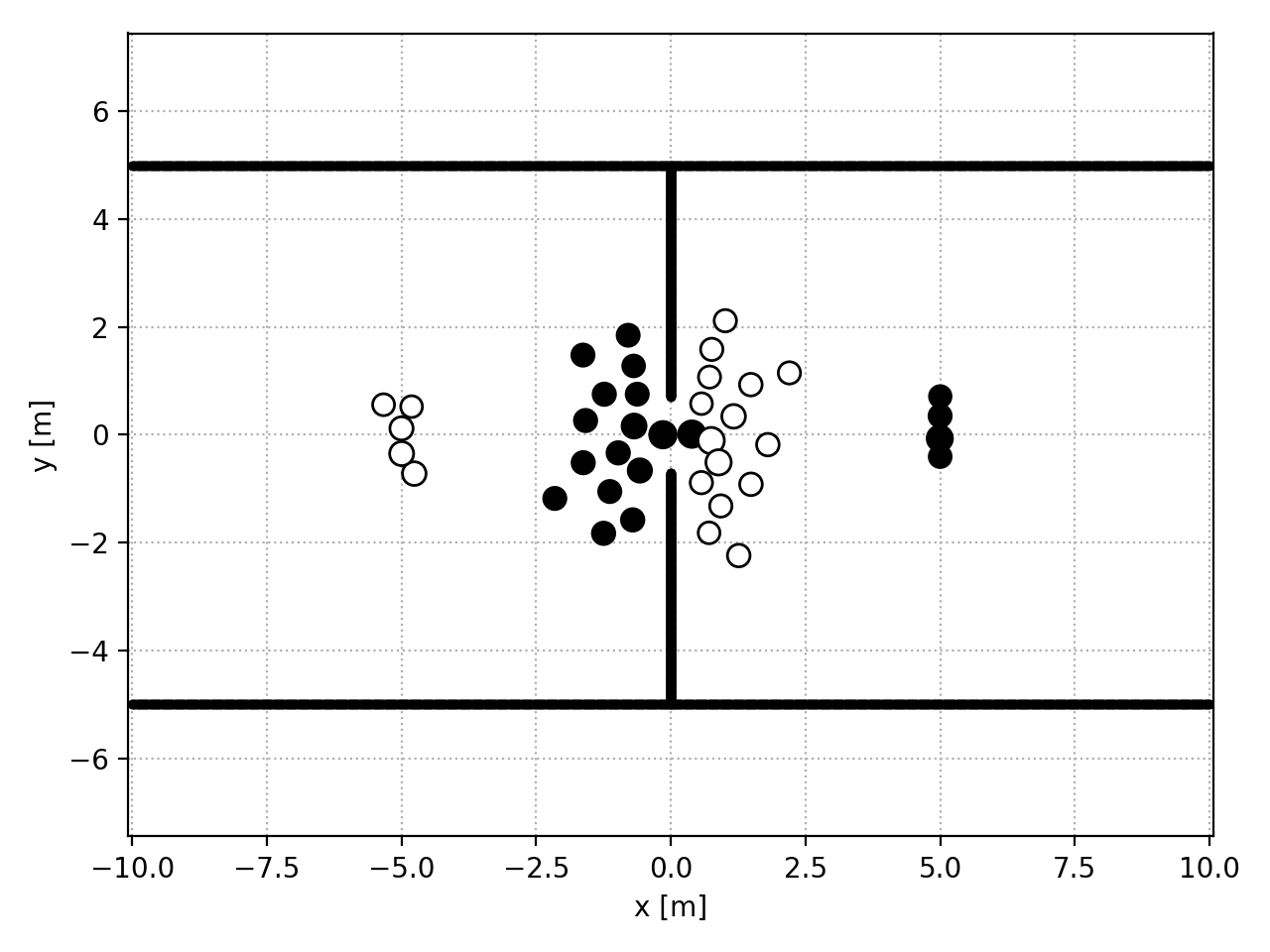}\\
  \caption{%
    Gate example.
    Pedestrians represented by filled circles walk towards a goal on the right
    and pedestrians represented by empty circles walk towards a goal on the left.
    Five pedestrians have already reached the left goal and four have reached
    the right goal. At this moment, a group of two pedestrians is passing
    towards the right.
    The size of the circle represents the walking speed of a pedestrian as
    in \cite{helbing1995social}.
  }
  \label{fig:sf-gate}
\end{figure}

\paragraph{Gate.}
Figure~\ref{fig:sf-gate} shows the gate example. Pedestrians are trying
to pass through a gate to get to the opposite side of the barrier.
In the original paper, the emergent behavior of grouping was observed:
groups of a few people pass the gate at a time. It was surprising that
pedestrians did not pass individually or that
not all of pedestrians of one side pass before the other side.
While I was able to reproduce that behavior, I want to point out that
this behavior is only present when the pedestrians have a preferred walking
speed distribution, \textit{i.e.}~pedestrians move at different preferred speeds.

\paragraph{Baseline Implementation.}
All operations are vectorized. The equations of motion are integrated
with the LeapFrog algorithm~\citep{birdsall2004plasma} with 25
integration steps per second.

\section{Method}

The goal is to create a more expressive Social Force model,
the \emph{Deep Social Force} model. The ultimate aim is to calibrate models
on large pedestrian datasets. The expressive detail of the model is adjustable
by the number of hidden units used for the deep neural networks and can increase
as more data becomes available in the future.
We have shown in our own work how a Social Force model can help a robot
navigate a crowded space~\citep{chen2019crowd}.
We have also
observed the limits of the Social Force model where two robots that need to pass
each other did not effectively negotiate the situation because they
had no preference to pass each other on the right. The aim is to discover these
intuitive behavioral patterns from data with this Deep Social Force model.

The technical implementation requires two key ingredients.
First, we use automatic
differentiation in PyTorch~\citep{paszke2017automatic}
to calculate the gradients of the interaction potentials to obtain the forces.
Second, we want to avoid imposing
assumptions on the potentials from their parametric form so we use
non-parametric potentials in the form of neural networks.

\paragraph{Automatic Differentiation of Potentials.}
With traditional methods, the dimensionality of the parameter space is crucial for
numerical optimizers and gradient approximation.
In addition, second-order optimizers like L-BFGS~\citep{byrd1995limited,broyden1970convergence,fletcher1970new,goldfarb1970family,shanno1970conditioning}
need to store an
approximation of the function curvature. Our proposed method provides exact
gradients without costly numerical gradient estimation and uses first-order
optimization methods -- Stochastic Gradient Descent~\citep{bottou2010large} --
that are suitable for the stochastic setting.
This is faster and
more stable than numeric differentiation and more flexible than hand-coded
derivatives of particular potentials.

\paragraph{Universal Potentials.}
Deep neural networks are universal function
approximators~\citep{cybenko1989approximation,hornik1989multilayer}
in certain limits.

A common parametric potential~\citep{helbing1995social} for human-human interactions
is a falling exponential:
\begin{equation}
  V_{\alpha\beta}(b) = V^0_{\alpha\beta} \exp(-b / \sigma)
\end{equation}
with $b$ being a function of the distance and velocities of the pedestrians and
where the parameters to optimize $\theta$ are the peak values of the
potential $V^0$ and width of the potential $\sigma$,
\textit{i.e.}~ $\theta = \{ V^0, \sigma \}$.
Here, a multi-layer perceptron (MLP) is used as a non-parametric%
\footnote{The term ``non-parametric'' refers to functions that can increase their capacity as more data becomes available; however, non-parametric functions do have parameters.}
replacement:
\begin{equation}
  V_{\alpha\beta}(b) = \textrm{Softplus} \;\; L_{1\times5} \;\; \textrm{Softplus} \;\; L_{5\times1} \;\; b
\end{equation}
where the parameters to optimize $\theta$ are the entries of the linear operators $\{ L_{1\times5}, L_{5\times1} \}$ and the chosen activation function and output non-linearity
here is a Softplus so that the outputs are bounded on one side only.
This MLP replacement removes all hand-engineered properties of parametric
potentials and discovers those and additional properties from data instead.

The potentials are not directly observed. The potential parameters $\theta$ are
inferred through the observed interactions of pedestrian paths. Every loss function
evaluation is a simulation of all pedestrian paths in the given scenes.
Automatic differentiation is used to obtain the gradients of the potentials.
It is the same technique that is used to obtain the gradients for the parameter
update with Stochastic Gradient Descent (SGD)~\citep{bottou2010large}.

\paragraph{Fourier Features.}
Recent progress in neural rendering has recaptured the research communities interest
in Fourier Features~\citep{rahimi2007random,tancik2020fourier}
as inputs to coordinate MLPs.
We construct multi-dimensional MLPs with Fourier Features (FF) that take the usual $b$
and in addition the perpendicular $d_{\perp}$ and parallel distance $d_{\parallel}$
as inputs. The first operation on these three inputs is a transformation to Fourier
Features. In particular, we use three one-dimensional feature transformations to 64
Fourier Features each. The complete MLP is:
\begin{equation}
  \textrm{MLP}(b, d_{\perp}, d_{\parallel})
  = \textrm{Softplus} \;\; L_{1\times64} \;\; \big[ \textrm{Softplus} \;\;
    L_{64\times64} \big]^3 \;\;  \textrm{Softplus} \;\; L_{64\times192} \;\;
    \textrm{FF}_{192 \times 3} \;\;
    \begin{bmatrix}
      b \\
      d_{\perp} \\
      d_{\parallel}
    \end{bmatrix}  \;\;.
\end{equation}

\section{Results}
\label{sec:results}

Fusing force-based models with deep neural networks is novel and
unconventional. The resulting model has the benefits of deep learning while still
being interpretable.


\paragraph{Training / Calibration.}
The aim is to train deep neural networks -- aka to calibrate the parameters
of interaction potentials -- with a method that scales to large
datasets. The deep learning community uses stochastic optimization methods
that operate on
batches of data and never on the whole dataset at once.
While traditional Social Force models are calibrated with second order
methods (\textit{e.g.}~\citet{yamaguchi2011you,seer2012kinects}),
we aim to use the first order Stochastic Gradient Descent (SGD)~\citep{bottou2010large}.
SGD can make use of the gradients that our
differentiable simulation provides and allows to operate on small batches of data.

\begin{figure}
  \centering
  \subfloat[circle scenes]{
    \includegraphics[height=5.7cm,trim=10cm 0 10cm 0,clip]{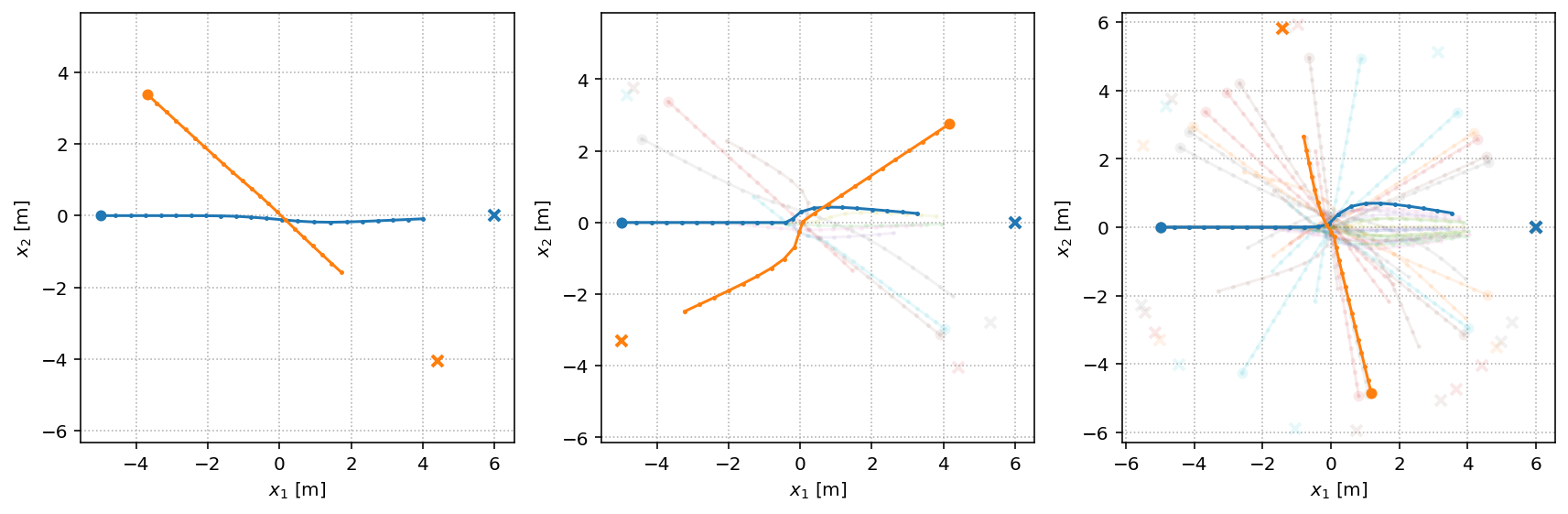}
    \label{fig:scenes-circle}
  }
  \subfloat[trajectory interactions]{
    \includegraphics[height=5.7cm,trim=0cm 0 0cm 0,clip]{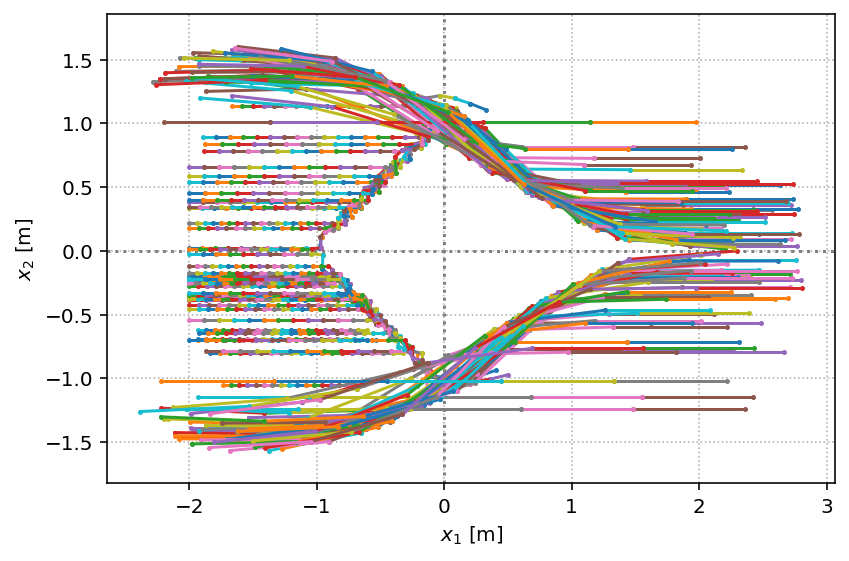}
    \label{fig:scenes-interaction}
  }
  \caption{
    In~(a), five ``circle scenes'' are shown and
    one is highlighted in blue and orange. The primary pedestrian (blue)
    always moves from left to right on the x-axis. The other pedestrian (orange)
    crosses the scene at a random angle.
    \href{https://www.svenkreiss.com/socialforce/scenarios.html}{\faBook~Notebook}.
    In~(b), the coordinate frame is fixed to the primary pedestrian and
    the trajectories of the secondary pedestrian are plotted for many synthetic
    scenes. In this particular instance, the interaction potential
    is shaped like a distorted diamond.
    \href{https://www.svenkreiss.com/socialforce/pedped_diamond.html\#scenarios}{\faBook~Notebook}.
  }
  \label{fig:scenes}
\end{figure}

\paragraph{Synthetic Scenes.}
A ``Scene'' is comprised of a few seconds of observed pedestrian trajectories,
see Figure~\ref{fig:scenes}.
Each trajectory is given
as a list of $(x, y)$ coordinates similar to latitudes and longitudes of GPS tracks.
We can use the interaction potentials given by \cite{helbing1995social}
or create our own to generate synthetic
scenes. The goal is to have a diverse set of these scenes that probe all aspects
of the interaction potential.

\paragraph{Automatic Differentiation of Potentials.}
With traditional methods, the dimensionality of the parameter space is crucial for
numerical optimizers and the first-order gradient approximation.
Automatic differentiation is faster and
more stable than numeric differentiation and more flexible than hand-coded
derivatives of particular potentials.
For a small multi-layer perceptron (MLP)
with only 10 parameters (1 input, 1 output, 5 hidden units),
it is possible to compare the numeric finite difference method
against back-propagation. Because of the
enormous reduction in function evaluations when back-propagated gradients are
available, the finite difference method is more than an order of magnitude slower
even on this small model.
Finite difference methods are infeasible for larger MLPs.
Below, we will see a parameter inference on large MLPs for 2D asymmetric potentials
with over 10,000 parameters using back-propagated gradients through
a differentiable simulation.

\begin{figure}
  \centering
  \includegraphics[height=5.3cm,trim=0 0 0cm 0,clip]{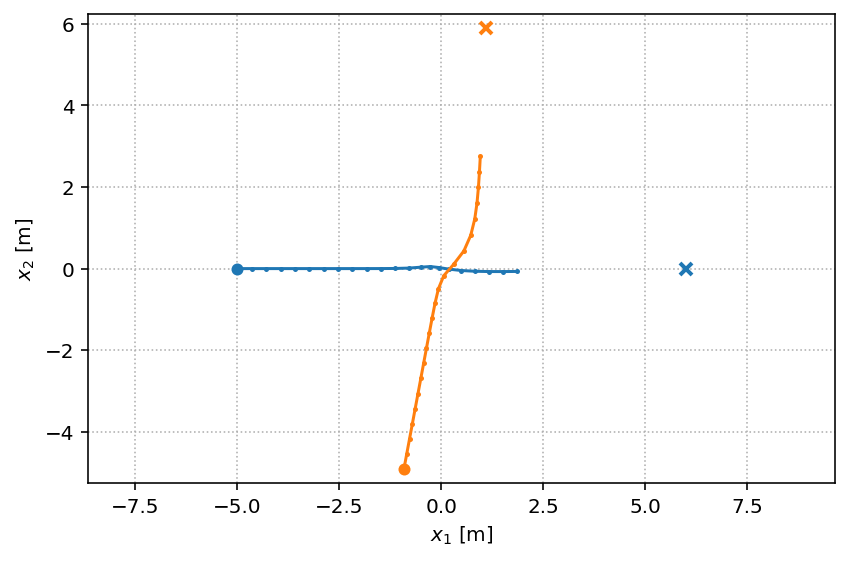}
  \includegraphics[height=5.3cm,trim=0 0 12.5cm 0,clip]{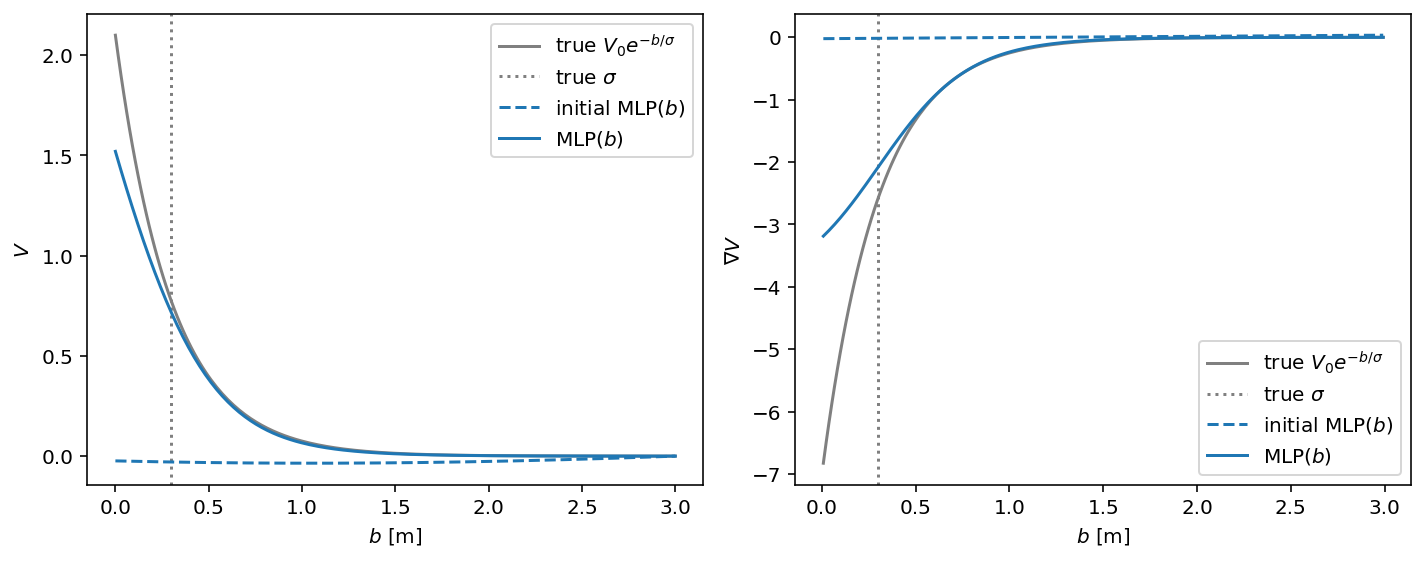}
  \caption{
    Fit of an MLP-based radial potential for human-human interaction
    to a single synthetic path of two pedestrians crossing and avoiding each other (left).
    The value $V$ is shown (right) as a function of reduced distance $b$.
    \href{https://www.svenkreiss.com/socialforce/pedped_1d.html}{\faBook~Notebook}.
  }
  \label{fig:mlp1d}
\end{figure}


\paragraph{Baseline: Parameter Inference for 1D MLP.}
The result of an inference of MLP parameters on synthetic data is shown
in Figure~\ref{fig:mlp1d}. This demonstrates that the differentiable simulation
can recover the exact potential as the generative potential with SGD.
Note that the inferred potential is a universal function in the form of
an MLP whereas the generating potential was a falling exponential function.


\begin{figure}
  \centering
  \subfloat[Social Force~\citep{helbing1995social}]{
    \includegraphics[width=0.48\textwidth,trim=0 0 15cm 0,clip]{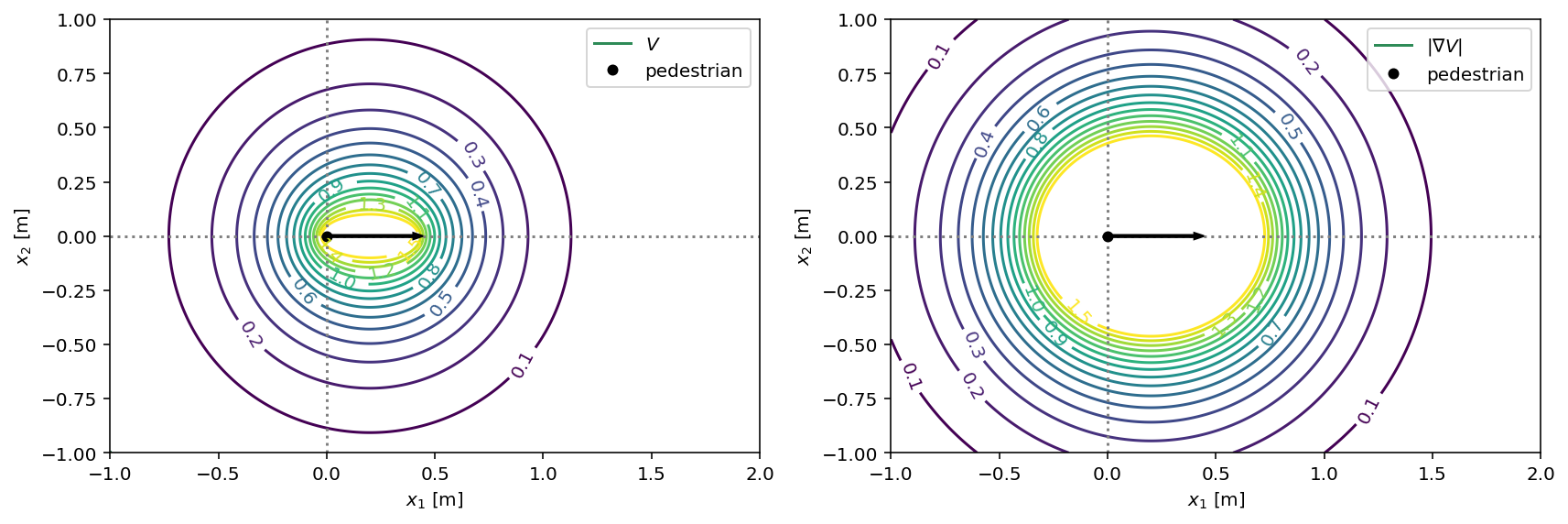}
  }
  \subfloat[Coordinate MLP with Fourier Features]{
    \includegraphics[width=0.48\textwidth,trim=0 0 15cm 0,clip]{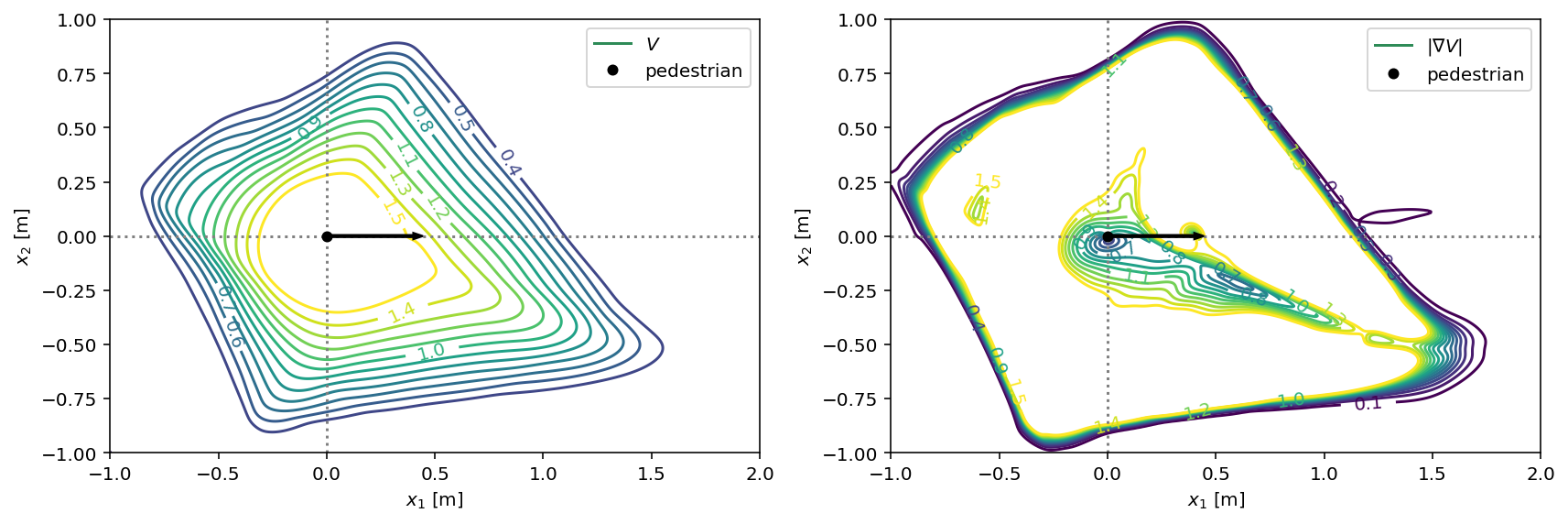}
  }
  \caption{
    A 2D visualization of the standard interaction potential in the original
    Social Force model~\citep{helbing1995social} is shown in (a).
    In (b), the result of an inference of a high-capacity MLP with about
    10,000 parameters on a synthetic simulation that was generated with an
    asymmetric interaction potential is shown.
    \href{https://www.svenkreiss.com/socialforce/pedped_diamond.html}{\faBook~Notebook}.
  }
  \label{fig:diamond-inference}
\end{figure}

\paragraph{Asymmetric 2D Potentials}
Above, the interaction potential in the one dimension of
reduced distance $b$~\citep{helbing1995social} was generalized with
a neural network. Now, we move to more generic 2D potentials.
Figure~\ref{fig:diamond-inference}a shows the interaction potential in 2D
for the classical Social Force model. It is an exact ellipse where the pedestrian
is at one of the focal points and the distance of the focal points of the
ellipse is related to the velocity of the pedestrian. The variable $b$
is the semi-minor axis of this ellipse.

To move to 2D, parallel $d_\parallel$
and perpendicular $d_\perp$ distances are added to the set of input variables.
However, I was not able to achieve any reasonable inference
results with this configuration. Inspired by recent successes in neural
rendering, I added a Fourier Feature
layer (FF)~\citep{tancik2020fourier,rahimi2007random}.
The FF layer takes the three input variables and converts them to 192
Fourier Features. The FF layer itself does not contain any trainable parameters.
The number of MLP parameters have nevertheless increased
to about 10,000 due to the increased input space and deeper architecture.

This 2D potential with its many parameters has a high capacity to model many
interaction types. With the differentiable simulation and gradient back-propagation,
we can optimize this large dimensional parameter space. On a simulation that
was generated with an asymmetric potential, we can infer that asymmetry
for this 2D MLP and obtain the potential that is shown in
Figure~\ref{fig:diamond-inference}b.

\begin{figure}
  \centering
  \includegraphics[width=0.8\linewidth,trim=13cm 0 0 0,clip]{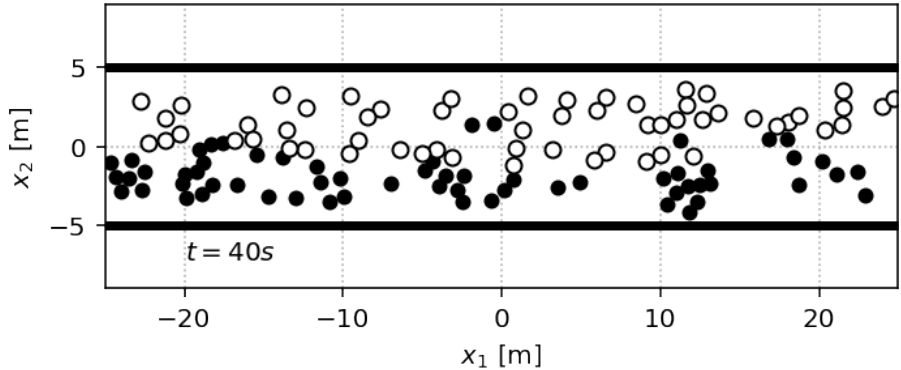}
  \caption{%
    Filled circles represent pedestrians that move to the right and
    empty circles pedestrians that move to the left.
    Because of the asymmetry of the new potential, pedestrians
    prefer to stay on the right side of oncoming pedestrians which implies
    that all pedestrians prefer the right side of the corridor in their
    direction of movement.
    \href{https://www.svenkreiss.com/socialforce/corridor.html\#asymmetric-diamond}{\faVideoCamera~Animation}.
    The corridor example is based on a scenario in \citet{helbing1995social}
    which is reproduced for reference in Figure~\ref{fig:sf-corridor}.
  }
  \label{fig:walkway-leftright-bias}
\end{figure}

We now have all ingredients to investigate implications from asymmetric potentials
in the ``corridor'' scenario. Figure~\ref{fig:walkway-leftright-bias} shows
a simulation of an asymmetric potential. It shows that
asymmetric potentials impose a left-right bias and can model people avoiding each
other on the right which ultimately leads to people preferring the right
side of the corridor. In this simulation, from random initializations, after 40s
all pedestrians have moved to the right in their walking direction as to avoid
other pedestrians.

\begin{figure*}
  \centering
  \subfloat[]{
    \includegraphics[height=6cm]{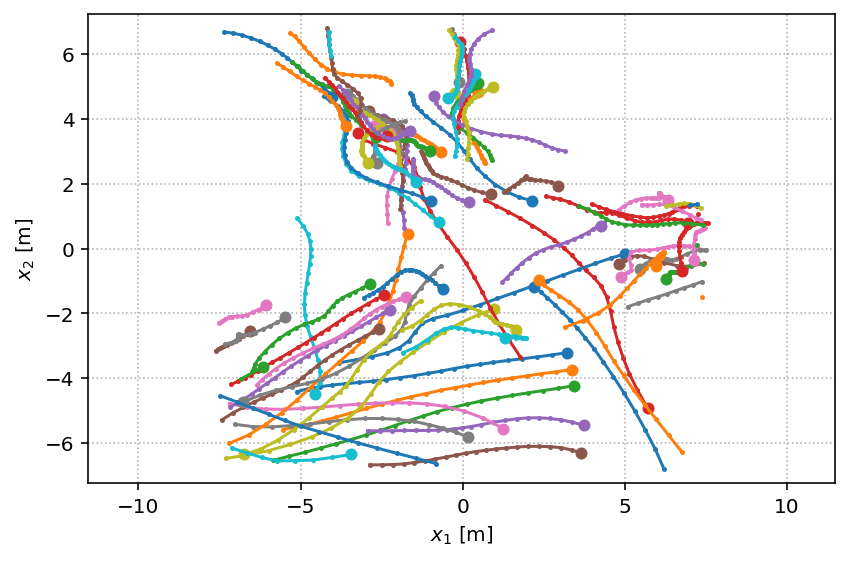}
    \label{fig:trajnet-scene}
  }
  \subfloat[]{
    \includegraphics[height=6cm,trim=0 0 7.5cm 0,clip]{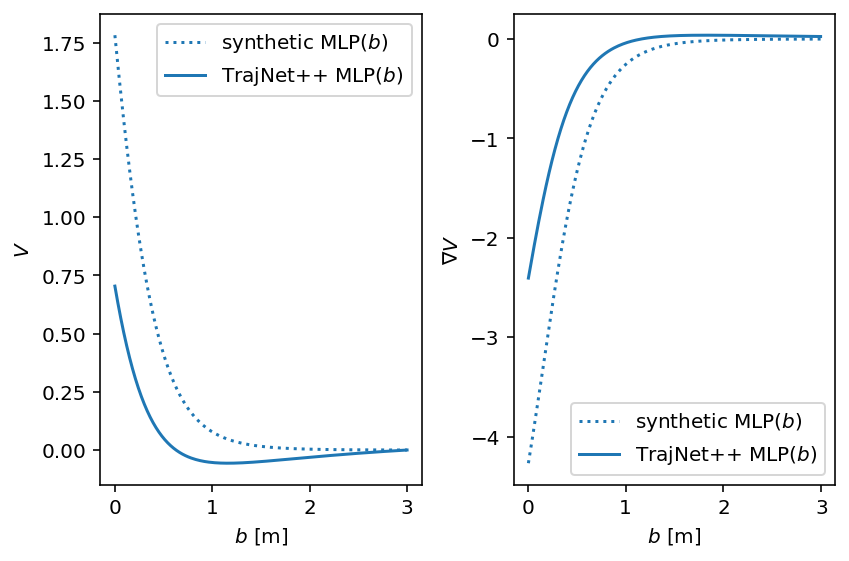}
    \label{fig:trajnet-inference-mlp1d}
  }
  \caption{
    In (a), a scene from the ``Crowd by Example'' dataset~\citep{Lerner2007CrowdsBE}
    as provided in the TrajNet++ challenge~\citep{kothari2020human} is shown.
    In (b), a first attempt at infering the shape of a 1D MLP
    potential on this real world dataset is shown.
    The MLP is initialized with synthetic data and then fine-tuned on the
    real world scene.
    \href{https://www.svenkreiss.com/socialforce/trajnet.html}{\faBook~Notebook}.
  }
  \label{fig:trajnet}
\end{figure*}

\paragraph{Inference from real data.}
Figure~\ref{fig:trajnet} shows a real world scenario on the left.
Infering an interaction potential from that scene is difficult.
On the right side of this figure, I share the result of a first attempt.
The MLP is initialized with a fit to synthetic
data and then fine-tuned on the real world example
from the ``Crowd by Example'' dataset~\citep{Lerner2007CrowdsBE}
as provided in the TrajNet++ challenge~\citep{kothari2020human}.
The optimization does not converge and does not work without the initialization
from synthetic data.
If one wanted to force an interpretation of this result, one would highlight the
two distinct regimes
below and above $b=1\,\textrm{m}$. For $b < 1\,\textrm{m}$, the potential is
repulsive as in the classical Social Force model. For $b > 1\,\textrm{m}$,
the potential is attractive. This type of behavior would have been impossible
to discover with a functional constraint to a falling exponential as in the
classical Social Force model~\citep{helbing1995social}. The reason for the
attractive nature of the potential is likely due to the large number of
social groups (groups of friends or classmates) in this dataset.
The classical Social Force
model actually does have a separate interaction term for the attractive forces
but here this behavior is discovered from data.

\section{Conclusion}

This paper and the associated open source software provide an entry point
to reproducible research.
The open source implementation allows to critically investigate previous
claims.

Most of this paper has focused on toy problems. It is important to
understand what works and what does not in simplified scenarios. However,
when applying these methods to real data, they generally disappoint.
Our modeling might still not be
advanced enough and we might need probabilistic path prediction and
evaluation methods. However, it might also be that the premise
of forecasting pedestrian paths from a few seconds of observations of
their $(x, y)$ coordinates is too optimistic for general situations.
This modeling assumption might be more valid for pedestrians
in crowded train station corridors where the main short-term motion is
driven by collision avoidance and less valid for the
motion of students across a sparsly populated open space on a
university campus where motion is driven by other factors not represented
in trajectory coordinates.

\acks{Sven Kreiss was hosted at the VITA lab at EPFL under
Prof.~Alexandre Alahi for the
duration of this project. This work was funded by the SNSF Spark
grant 190677.}

\vskip 0.2in
\bibliography{references}

\end{document}